\documentclass[runningheads]{llncs}
\usepackage{graphicx}
\usepackage{comment}
\usepackage{amsmath,amssymb} 
\usepackage{color}
\usepackage{multirow}
    
\newcommand{\norm}[1]{\left\lVert#1\right\rVert}

\begin{document}
\pagestyle{headings}
\mainmatter
\def\ECCVSubNumber{2950}  

\title{PointPWC-Net: Cost Volume on Point Clouds for (Self-)Supervised Scene Flow Estimation} 

\titlerunning{PointPWC-Net}
\author{Wenxuan Wu\inst{1} \and
Zhi Yuan Wang \inst{2} \and
Zhuwen Li\inst{2} \and 
Wei Liu\inst{2} \and
Li Fuxin\inst{1}}
\authorrunning{W. Wu et al.}
%
\institute{CORIS Institute, Oregon State University\\
\email{\{wuwen,lif\}@oregonstate.edu} \and
Nuro, Inc.}
\maketitle

\begin{abstract}
We propose a novel end-to-end deep scene flow model, called PointPWC-Net, that directly processes 3D point cloud scenes with large motions in a coarse-to-fine fashion. Flow computed at the coarse level is upsampled and warped to a finer level, enabling the algorithm to accommodate for large motion without a prohibitive search space. We introduce novel cost volume, upsampling, and warping layers to efficiently handle 3D point cloud data. Unlike traditional cost volumes that require exhaustively computing all the cost values on a high-dimensional grid, our point-based formulation discretizes the cost volume onto input 3D points, and a PointConv operation efficiently computes convolutions on the cost volume. Experiment results on FlyingThings3D and KITTI outperform the state-of-the-art by a large margin. We further explore novel self-supervised losses to train our model and achieve comparable results to state-of-the-art trained with supervised loss. Without any fine-tuning, our method also shows great generalization ability on the KITTI Scene Flow 2015 dataset, outperforming all previous methods. The code is released at {\color{red}\url{https://github.com/DylanWusee/PointPWC}}.
\end{abstract}

\section{Introduction}
Scene flow is the 3D displacement vector between each surface point in two consecutive frames. As a fundamental tool for low-level understanding of the world, scene flow can be used in many 3D applications including autonomous driving.
Traditionally, scene flow was estimated directly from RGB data~\cite{menze2015object,mayer2016large,vedula1999three,vogel2013piecewise}. But recently, due to the increasing application of 3D sensors such as LiDAR, there is interest on directly estimating scene flow from 3D point clouds. 

Fueled by recent advances in 3D deep networks that learn effective feature representations directly from point cloud data, recent work adopt ideas from 2D deep optical flow networks to 3D to estimate scene flow from point clouds. 
FlowNet3D~\cite{liu2019flownet3d} operates directly on points with PointNet++~\cite{qi2017pointnet++}, and proposes a \textit{flow embedding} which is computed in one layer to capture the correlation between two point clouds, and then propagates it through finer layers to estimate the scene flow. HPLFlowNet~\cite{gu2019hplflownet} computes the correlation jointly from multiple scales utilizing the upsampling operation in bilateral convolutional layers.


An important piece in deep optical flow estimation networks is the cost volume~\cite{kendall2017end,xu2017accurate,sun2018pwc}, a 3D tensor that contains matching information between neighboring pixel pairs from consecutive frames. 
In this paper, we propose a novel learnable point-based cost volume where we discretize the cost volume to input point pairs, avoiding the creation of a dense 4D tensor if we naively extend from the image to point cloud. Then we apply the efficient PointConv layer~\cite{wu2019pointconv} on this irregularly discretized cost volume. We experimentally show that it outperforms previous approaches for associating point cloud correspondences, as well as the cost volume used in 2D optical flow. We also propose efficient upsampling and warping layers to implement a coarse-to-fine flow estimation framework.


As in optical flow, it is difficult and expensive to acquire accurate scene flow labels for point clouds. Hence, beyond supervised scene flow estimation, we also explore self-supervised scene flow which does not require human annotations. 
We propose new self-supervised loss terms: Chamfer distance~\cite{fan2017point}, smoothness constraint and Laplacian regularization. These loss terms enable us to achieve state-of-the-art performance without any 
supervision. 

\begin{figure}[t]
    \centering
    \includegraphics[width=0.8\textwidth]{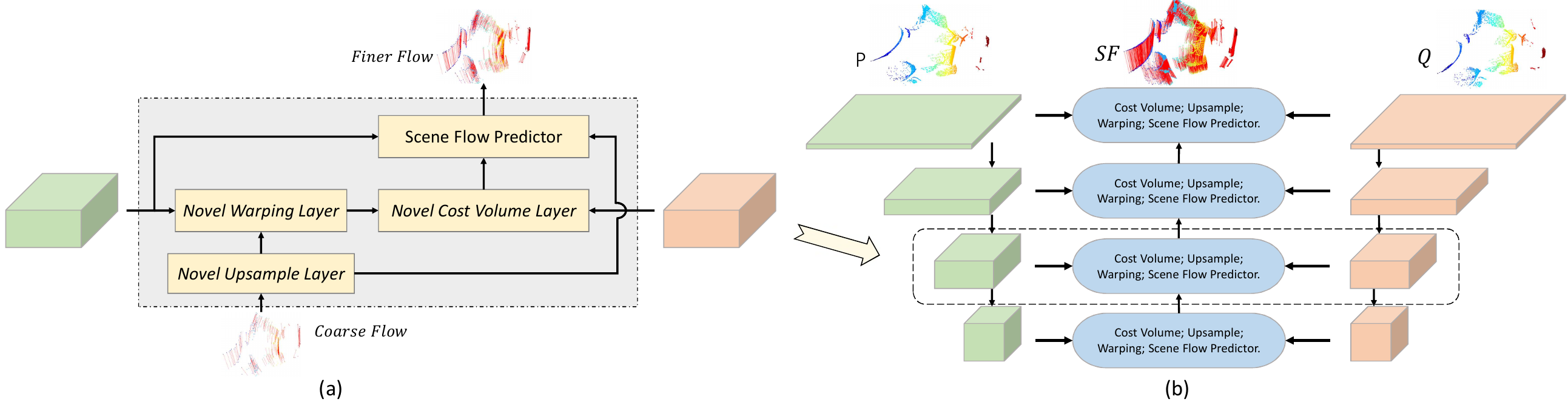}
    \caption{(a) illustrates how the pyramid features are used by the novel cost volume, warping, and upsampling layers in one level. (b) shows the overview structure of PointPWC-Net. At each level, PointPWC-Net first warps features from the first point cloud using the upsampled scene flow. Then, the cost volume is computed using features from the warped first point cloud and the second point cloud. Finally, the scene flow predictor predicts finer flow at the current level using features from the first point cloud, the cost volume, and the upsampled flow. (Best viewed in color)}
    \label{fig:PointPWCNet}
\end{figure}

We conduct extensive experiments on FlyingThings3D~\cite{mayer2016large} and KITTI Scene Flow 2015~\cite{Menze2018JPRS,Menze2015ISA} datasets with both supervised loss and the proposed self-supervised losses. Experiments show that the proposed PointPWC-Net outperforms all previous methods by a large margin. The self-supervised version is comparable with some of the previous supervised methods on FlyingThings3D, such as SPLATFlowNet~\cite{su2018splatnet}. On KITTI where supervision is not available, our self-supervised version achieves better performance than the supervised version trained on FlyingThings3D, far surpassing state-of-the-art. We also ablate each critical component of PointPWC-Net to understand their contributions.

The key contributions of our work are: 
\\ \indent $\bullet$ We propose a novel learnable cost volume layer that performs convolution on the cost volume without creating a dense 4D tensor. 
\\ \indent $\bullet$ With the novel learnable cost volume layer, we present a novel model, called PointPWC-Net, that estimates scene flow from two consecutive point clouds in a coarse-to-fine fashion.
\\ \indent $\bullet$ We introduce self-supervised losses that can train PointPWC-Net without any ground truth label. To our knowledge, we are among the first to propose such an idea in 3D point cloud deep scene flow estimation.
\\ \indent $\bullet$ We achieve state-of-the-art performance on FlyingThing3D and KITTI Scene Flow 2015, far surpassing previous state-of-the-art.

\section{Related Work}

\noindent \textbf{Deep Learning on Point Clouds.} Deep learning methods on 3D point clouds have gained more attention in the past several years. Some latest work~\cite{ravanbakhsh2016deep,qi2017pointnet,qi2017pointnet++,su2018splatnet,tatarchenko2018tangent,hua2018pointwise,groh2018flex,verma2018feastnet,li2018pointcnn} directly take raw point clouds as input. \cite{ravanbakhsh2016deep,qi2017pointnet,qi2017pointnet++} use a shared multi-layer perceptron (MLP) and max pooling layer to obtain features of point clouds. 
Other work~\cite{simonovsky2017dynamic,jia2016dynamic,wang2019dynamic,hermosilla2018monte,wang2018deep,wu2019pointconv} propose to learn continuous convolutional filter weights as a nonlinear function from 3D point coordinates, approximated with MLP. 
\cite{hermosilla2018monte,wu2019pointconv} use a density estimation to compensate the non-uniform sampling, and \cite{wu2019pointconv} significantly improves the memory efficiency by a change of summation trick, allowing these networks to scale up and achieving comparable capabilities with 2D convolution.

\noindent \textbf{Optical Flow Estimation.} Optical flow estimation is a core computer vision problem and has many applications. Traditionally, top performing methods often adopt the energy minimization approach ~\cite{horn1981determining} and a coarse-to-fine, warping-based method~\cite{bergen1992hierarchical,bruhn2005lucas,brox2004high}. Since FlowNet~\cite{dosovitskiy2015flownet}, there were many recent work using a deep network to learn optical flow. \cite{ilg2017flownet} stacks several FlowNets into a larger one. \cite{ranjan2017optical} develops a compact spatial pyramid network. \cite{sun2018pwc} integrates the widely used traditional pyramid, warping, and cost volume technique into CNNs for optical flow, and outperform all the previous methods with high efficiency. We utilized a basic structure similar to theirs but proposed novel cost volume, warping and upsampling layers appropriate for point clouds. 

\noindent \textbf{Scene Flow Estimation.} 3D scene flow is first introduced by \cite{vedula1999three}. Many works~\cite{huguet2007variational,menze2015object,vogel20153d} estimate scene flow using RGB data. \cite{huguet2007variational} introduces a variational method to estimate scene flow from stereo sequences. \cite{menze2015object} proposes an object-level scene flow estimation approach and introduces a dataset for 3D scene flow. \cite{vogel20153d} presents a piecewise rigid scene model for 3D scene flow estimation. 

Recently, there are some works~\cite{dewan2016rigid,ushani2017learning,ushani2018feature} that estimate scene flow directly from point clouds using classical techniques. \cite{dewan2016rigid} introduces a method that formulates the scene flow estimation problem as an energy minimization problem with assumptions on local geometric constancy and regularization for motion smoothness. \cite{ushani2017learning} proposes a real-time four-steps method of constructing occupancy grids, filtering the background, solving an energy minimization problem, and refining with a filtering framework. \cite{ushani2018feature} further improves the method in \cite{ushani2017learning} by using an encoding network to learn features from an occupancy grid. 

In some most recent work~\cite{wang2018deep,liu2019flownet3d,gu2019hplflownet}, researchers attempt to estimate scene flow from point clouds using deep learning in a end-to-end fashion. \cite{wang2018deep} uses PCNN to operate on LiDAR data to estimate LiDAR motion. \cite{liu2019flownet3d} introduces FlowNet3D based on PointNet++~\cite{qi2017pointnet++}. FlowNet3D uses a flow embedding layer to encode the motion of point clouds. However, it requires encoding a large neighborhood in order to capture large motions. \cite{gu2019hplflownet} presents HPLFlowNet to estimate the scene flow using Bilateral Convolutional Layers(BCL), which projects the point cloud onto a permutohedral lattice. \cite{behl2019pointflownet} estimates scene flow with a network that jointly predicts 3D bounding boxes and rigid motions of objects or background in the scene. Different from \cite{behl2019pointflownet}, we do not require the rigid motion assumption and segmentation level supervision to estimate scene flow.

\noindent \textbf{Self-supervised Scene Flow.} There are several recent works~\cite{liu2019unsupervised,yin2018geonet,zou2018df,lee2019cemnet,hur2020self} which jointly estimate multiple tasks, i.e. depth, optical flow, ego-motion and camera pose without supervision. They take 2D images as input, which have ambiguity when used in scene flow estimation. In this paper, we investigate self-supervised learning of scene flow from 3D point clouds with our PointPWC-Net. Concurrently, Mittal \textit{et al.}~\cite{mittal2020just} introduced Nearest Neighbor(NN) Loss and Cycle Consistency Loss to self-supervised scene flow estimation from point clouds. However, they does not take the local structure properties of 3D point clouds into consideration. In our work, we propose to use smoothness and Laplacian coordinates to preserve local structure for scene flow.

\noindent \textbf{Traditional Point Cloud Registration.} Point cloud registration has been extensively studied well before deep learning~\cite{guo20143d,tam2012registration}. Most of the work~\cite{choi2015robust,gelfand2005robust,holz2015registration,makadia2006fully,mian2006three,shin2010unsupervised,theiler2015globally,zhou2016fast} only works when most of the motion in the scene is globally rigid. Many methods are based on the iterative closest point(ICP)~\cite{besl1992method} and its variants~\cite{pomerleau2013comparing}. 
Several works~\cite{amberg2007optimal,myronenko2010point,jain2007non,brown2007global} deal with non-rigid point cloud registration. Coherent Point Drift(CPD)~\cite{myronenko2010point} introduces a probabilistic method for both rigid and non-rigid point set registration. However, the computation overhead makes it hard to apply on real world data in real-time. Many algorithms are proposed to extend the CPD method~\cite{ma2015non,nguyen2015multiple,ma2014mixture,zhou2014robust,ge2014non,ma2015robust,fu2016non,bai2017nonrigid,ge2015non,ma2017non,ma2018guided,lei2017fast,tao2014asymmetrical,saval20183d,danelljan2016probabilistic,lu2015accelerated,zhang2018non,wang2017fuzzy,yu2015fast,qu2016probabilistic}. 
Some algorithms require additional information for point set registration. The work~\cite{saval20183d,danelljan2016probabilistic} takes the color information along with the spatial location into account. \cite{amberg2007optimal} requires meshes for non-rigid registration. In~\cite{qu2016probabilistic}, the regression and clustering for point set registration in a Bayesian framework are presented. All the aforementioned work require optimization at inference time, which has significantly higher computation cost than our method which run in a fraction of a second during inference. 


\section{Approach}\label{sec:pointpwcnet}
To compute optical flow with high accuracy, one of the most important components is the cost volume. In 2D images, the cost volume can be computed by aggregating the cost in a square neighborhood on a grid. However, computing cost volume across two point clouds is difficult since 3D point clouds are unordered with a nonuniform sampling density. In this section, we introduce a novel learnable cost volume layer, and use it to construct a deep network with the help of other auxiliary layers that outputs high quality scene flow.

\subsection{The Cost Volume Layer}

As one of the key components of optical flow estimation, most state-of-the-art algorithms, both traditional~\cite{sun2014quantitative,revaud2015epicflow} and modern deep learning based ones~\cite{sun2018pwc,xu2017accurate,chabra2019stereodrnet}, use the cost volume to estimate optical flow. However, computing cost volumes on point clouds is still an open problem. There are several works~\cite{liu2019flownet3d,gu2019hplflownet} that compute some kind of flow embedding or correlation between point clouds. \cite{liu2019flownet3d} proposes a flow embedding layer to aggregate feature similarities and spatial relationships to encode point motions. However, the motion information between points can be lost due to the max pooling operation in the flow embedding layering. \cite{gu2019hplflownet} introduces a CorrBCL layer to compute the correlation between two point clouds, which requires to transfer two point clouds onto the same permutohedral lattice. 

To address these issues, we present a novel learnable cost volume layer directly on the features of two point clouds.
Suppose $f_i \in \mathbb{R}^c$ is the feature for point $p_i \in P$ and $g_j \in \mathbb{R}^c$ the feature for point $q_j \in Q$, the matching cost between $p_i$ and $q_j$ can be defined as:
\begin{align}\label{eq:matchingcost}
    Cost(p_i, q_j) &= h(f_i, g_j, q_j, p_i) \\
                   &= MLP(concat(f_i, g_j, q_j - p_i))
\end{align}
Where \textit{concat} stands for concatenation.
In our network, the feature $f_i$ and $g_j$ are either the raw coordinates of the point clouds, or the convolution output from previous layers. The intuition is that, as a universal approximator, MLP should be able to learn the potentially nonlinear relationship between the two points. Due to the flexibility of the point cloud, we also add a direction vector $(q_j - p_i)$ to the computation besides the point features $f_i$ and $g_j$. %

Once we have the matching costs, they can be aggregated as a cost volume for predicting the movement between two point clouds. In 2D images, aggregating the cost is simply by applying some convolutional layers as in PWC-Net~\cite{sun2018pwc}. However, traditional convolutional layers can not be applied directly on point clouds due to their unorderness. \cite{liu2019flownet3d} uses max-pooing to aggregate features in the second point cloud. \cite{gu2019hplflownet} uses CorrBCL to aggregate features on a permutohedral lattice. However, their methods only aggregate costs in a point-to-point manner, which is sensitive to outliers. To obtain robust and stable cost volumes, in this work, we propose to aggregate costs in a patch-to-patch manner similar to the cost volumes on 2D images~\cite{kendall2017end,sun2018pwc}. 

\begin{figure}[t]
    \centering
    \includegraphics[width=0.8\textwidth]{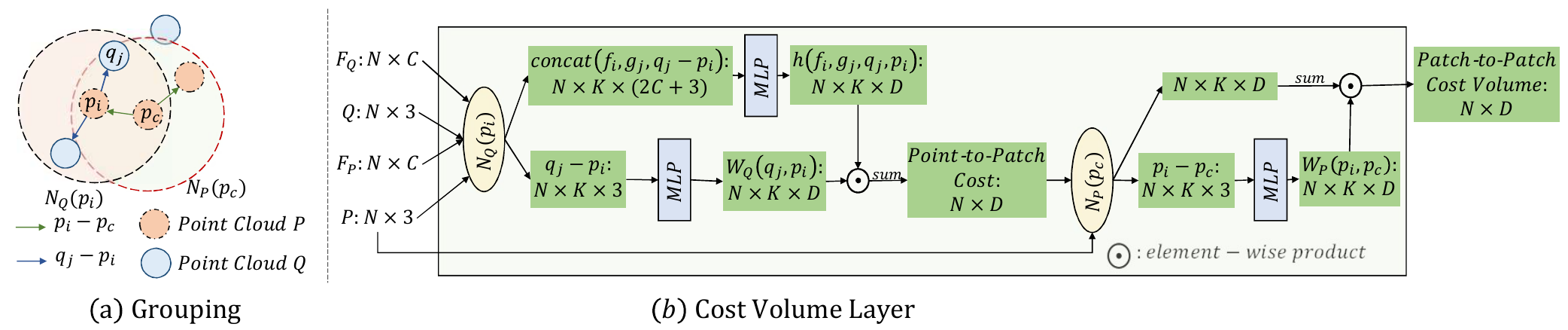}
    \caption{(a)~\textbf{Grouping.} For a point $p_c$, we form its $K$-NN neighborhoods in each point cloud as $N_P(p_c)$ and $N_Q(p_c)$ for cost volume aggregation. We first aggregate the cost from the patch $N_Q(p_c)$ in point cloud $Q$. Then, we aggregate the cost from patch $N_P(p_c)$ in the point cloud $P$. (b)~\textbf{Cost Volume Layer.} The features of neighboring points in $N_Q(p_c)$ are concatenated with the direction vector $(q_i-p_c)$ to learn a point-to-patch cost between $p_c$ and $Q$ with PointConv. Then the point-to-patch costs in $N_P(p_c)$ are further  aggregated with PointConv to construct a patch-to-patch cost volume}
    \label{fig:group_cost}
\end{figure}

For a point $p_c$ in $P$, we first find a neighborhood $N_P(p_c)$ around $p_c$ in $P$. For each point $p_i \in N_P(p_c)$, we find a neighborhood $N_Q(p_i)$ around $p_i$ in $Q$. The cost volume for $p_c$ is defined as:
\begin{align}\label{eq:costvolume}\small
    \tiny CV(p_c) &= \sum_{p_i \in N_P(p_c)} W_P(p_i,p_c) \sum_{q_j \in N_Q(p_i)} W_Q(q_j,p_i) ~cost(q_j, p_i) \\
    W_P(p_i,  p_c) &= MLP(p_i - p_c) \\
    W_Q(q_j, p_i) &= MLP(q_j - p_i)
\end{align}
Where $W_P(p_i, p_c)$ and $W_Q(q_j, p_i)$ are the convolutional weights \textit{w.r.t} the direction vectors that are used to aggregate the costs from the patches in $P$ and $Q$. It is learned as a continuous function of the directional vectors $(q_i - p_c) \in \mathbb{R}^3$ and $(q_j - p_i) \in \mathbb{R}^3$, respectively with an MLP, as in~\cite{wu2019pointconv} and PCNN~\cite{wang2018deep}. The output of the cost volume layer is a tensor with shape $(n_1, D)$, where $n_1$ is the number of points in $P$, and $D$ is the dimension of the cost volume, which encodes all the motion information for each point. The patch-to-patch idea used in the cost volume is illustrated in Fig.~\ref{fig:group_cost}.

There are two major differences between this cost volume for scene flow of 3D point clouds and conventional 2D cost volumes for stereo and optical flow. The first one is that we introduce a learnable function $cost(\cdot)$ that can dynamically learn the cost or correlation within the point cloud structures. Ablation studies in Sec.\ref{sec:ablation} show that this novel learnable design achieve better results than traditional cost volume~\cite{sun2018pwc} in scene flow estimation. The second one is that this cost volume is discretized irregularly on the two input point clouds and their costs are aggregated with point-based convolution. Previously, in order to compute the cost volume for optical flow in a $d \times d$ area on a $W \times H$ 2D image, all the values in a $d^2 \times W \times H$ tensor needs to be populated, which is already slow to compute in 2D, but would be prohibitively costly in the 3D space. With (volumetric) 3D convolution, one needs to search a $d^3$ area to obtain a cost volume in 3D space. Our cost volume discretizes on input points and avoids this costly operation, while essentially creating the same capabilities to perform convolutions on the cost volume. With the proposed cost volume layer, we only need to find two neighborhoods $N_P(p_c)$ and $N_Q(p_i)$ of size $K$, which is much cheaper and does not depend on the number of points in a point cloud. In our experiments, we fix $|N_P(p_c)| = |N_Q(p_i)| = 16$. If a larger neighborhood is needed, we could subsample the neighborhood which would bring it back to the same speed. This subsampling operation is only applicable to the sparse point cloud convolution and not possible for conventional volumetric convolutions. We anticipate this novel cost volume layer to be widely useful beyond scene flow estimation. Table~\ref{tab:modeldesign} shows that it is better than \cite{liu2019flownet3d}'s MLP+Maxpool strategy.

\subsection{PointPWC-Net}
Given the proposed learnable cost volume layer, we construct a deep network for scene flow estimation. As demonstrated in 2D optical flow estimation, one of the most effective methods for dense estimation is the coarse-to-fine structure. In this section, we introduce some novel auxiliary layers for point clouds that construct a coarse-to-fine network for scene flow estimation along with the proposed learnable cost volume layer. The network is called \textit{``PointPWC-Net''} following~\cite{sun2018pwc}.

As shown in Fig.\ref{fig:PointPWCNet}, PointPWC-Net predicts dense scene flow in a coarse-to-fine fashion. The input to PointPWC-Net is two consecutive point clouds, $P=\{p_i \in \mathbb{R}^3\}^{n_1}_{i = 1}$ with $n_1$ points, and $Q=\{q_j \in \mathbb{R}^3\}^{n_2}_{j = 1}$ with $n_2$ points. We first construct a feature pyramid for each point cloud. Afterwards, we build a cost volume using features from both point clouds at each layer. Then, we use the feature from $P$, the cost volume, and the upsampled flow to estimate the finer scene flow. We take the predicted scene flow as the coarse flow, upsample it to a finer flow, and warp points from $P$ onto $Q$. Note that both the upsampling and the warping layers are efficient with no learnable parameters.

\noindent \textbf{Feature Pyramid from Point Cloud.} To estimate scene flow with high accuracy, we need to extract strong features from the input point clouds. 
 We generate an $L$-level pyramid of feature representations, with the top level being the input point clouds, i.e., $l_0 = P/Q$. For each level $l$, we use furthest point sampling~\cite{qi2017pointnet++} to downsample the points by a factor of 4 from previous level $l - 1$, and use PointConv~\cite{wu2019pointconv} to perform convolution on the features from level $l - 1$. As a result, we can generate a feature pyramid with $L$ levels for each input point cloud. After this, we enlarge the receptive field at level $l$ of the pyramid by upsampling the feature in level $l+1$ 
and concatenate it to the feature at level $l$. 

\noindent \textbf{Upsampling Layer.} The upsampling layer can propagate the scene flow estimated from a coarse layer to a finer layer. We use a distance based interpolation to upsample the coarse flow. Let $P^l$ be the point cloud at level $l$, $SF^l$ be the estimated scene flow at level $l$, and $p^{l-1}$ be the point cloud at level $l-1$. For each point $p^{l-1}_i$ in the finer level point cloud $P^{l - 1}$, we can find its K nearest neighbors $N(p^{l-1}_i)$ in its coarser level point cloud $P^{l}$. The interpolated scene flow of finer level $SF^{l-1}$ is computed using inverse distance weighted interpolation:
\begin{equation}\label{eq:interpolation}\small
    SF^{l-1}(p_i) = \frac{\sum^k_{j=1}w(p^{l-1}_i, p^l_j)SF^l(p^l_j)}{\sum^k_{j=1}w(p^{l-1}_i, p^l_j)}
\end{equation}

\noindent where $w(p^{l-1}_i, p^l_j) = 1/d(p^{l-1}_i, p^l_j)$, $p^{l-1}_i \in P^{l-1}$, and $p^l_j \in N(p^{l-1}_i)$. $d(p^{l-1}_i, p^l_j)$ is a distance metric. We use Euclidean distance in this work.
    
\noindent \textbf{Warping Layer.} Warping would ``apply" the computed flow so that  only the residual flow needs to be estimated afterwards, hence the search radius can be smaller when constructing the cost volume. In our network, we first up-sample the scene flow from the previous coarser level and then warp it before computing the cost volume.
Denote the upsampled scene flow as $SF=\{sf_i \in \mathbb{R}^3\}^{n_1}_{i = 1}$, and the warped point cloud as $P_w=\{p_{w,i} \in \mathbb{R}^3\}^{n_1}_{i = 1}$. The warping layer is simply an element-wise addition between the upsampled and computed scene flow $P_w = \{p_{w,i} = p_i + sf_i | p_i \in P, sf_i \in SF \}^{n_1}_{i = 1}$.
A similar warping operation is used for visualization to compare the estimated flow with the ground truth in~\cite{liu2019flownet3d,gu2019hplflownet}, but not used in coarse-to-fine estimation. \cite{gu2019hplflownet} uses an offset strategy to reduce search radius which is specific to the permutohedral lattice. 

\noindent \textbf{Scene Flow Predictor.} In order to obtain a flow estimate at each level, a convolutional scene flow predictor is built as multiple layers of PointConv and MLP. The inputs of the flow predictor are the cost volume, the feature of the first point cloud, the up-sampled flow from previous layer and the up-sampled feature of the second last layer from previous level's scene flow predictor, which we call the predictor feature. 
The output is the scene flow $SF=\{sf_i \in \mathbb{R}^3\}^{n_1}_{i=1}$ of the first point cloud $P$. The first several PointConv layers are used to merge the feature locally, and the following MLP is used to estimate the scene flow on each point. We keep the flow predictor structure at different levels the same, but the parameters are not shared.

\section{Training Loss Functions}
In this section, we introduce two loss functions to train PointPWC-Net for scene flow estimation. One is the standard multi-scale supervised training loss, which has been explored in deep optical flow estimation~\cite{sun2018pwc} in 2D images. We use this supervised loss to train the model for fair comparison with previous scene flow estimation work, including FlowNet3D~\cite{liu2019flownet3d} and HPLFlowNet~\cite{gu2019hplflownet}. Due to that acquiring densely labeled 3D scene flow dataset is extremely hard, we also propose a novel self-supervised loss to train our PointPWC-Net without any supervision.

\subsection{Supervised Loss}
We adopt the multi-scale loss function in FlowNet~\cite{dosovitskiy2015flownet} and PWC-Net~\cite{sun2018pwc} as a supervised learning loss to demonstrate the effectiveness of the network structure and the design choice. Let $SF^l_{GT}$ be the ground truth flow at the $l$-th level. The multi-scale training loss $\boldsymbol\ell(\Theta) = \sum^{L}_{l=l_0}\alpha_l\sum_{p \in P} \norm{SF^l_{\Theta}(p) - SF^l_{GT}(p)}_2$ is used where $\norm{\cdot}_2$ computes the $L_2$-norm, $\alpha_l$ is the weight for each pyramid level $l$, 
and $\Theta$ is the set of all the learnable parameters in our PointPWC-Net, including the feature extractor, cost volume layer and scene flow predictor at different pyramid levels. Note that the flow loss is not squared as in~\cite{sun2018pwc} for robustness.

\subsection{Self-supervised Loss}
Obtaining the ground truth scene flow for 3D point clouds is difficult and there are not many publicly available datasets for scene flow learning from point clouds. In this section, we propose a self-supervised learning objective function to learn the scene flow in 3D point clouds without supervision. Our loss function contains three parts: \textit{Chamfer distance}, \textit{Smoothness constraint}, and \textit{Laplacian regularization}~\cite{wang2018pixel2mesh,sorkine2005laplacian}. To the best of our knowledge, we are the first to study self-supervised deep learning of scene flow estimation from 3D point clouds, concurrent with ~\cite{mittal2020just}.

\noindent \textbf{Chamfer Loss} The goal of Chamfer loss is to estimate scene flow by moving the first point cloud as close as the second one. Let $SF^l_{\Theta}$ be the scene flow predicted at level $l$. Let $P^l_{w}$ be the point cloud warped from the first point cloud $P^l$ according to $SF^l_{\Theta}$ in level $l$, $Q^l$ be the second point cloud at level $l$. Let $p^l_{w}$ and $q^l$ be points in $P^l_{w}$ and $Q^l$. The Chamfer loss $\ell^l_C$ can be written as:
\begin{eqnarray}\label{eq:chamfer}
    \small
    P^l_w  & = & P^l + SF^l_{\Theta}\\
    \ell^l_C(P^l_{w}, Q^l) & = & \sum_{p^l_w\in P^l_{w}} \min_{q^l \in Q^l} \norm{p^l_w - q^l}^2_2 + \sum_{q^l\in Q^l} \min_{p^l_w\in P^l_{w}} \norm{p^l_w - q^l}^2_2 \nonumber
\end{eqnarray}

\noindent \textbf{Smoothness Constraint} In order to enforce local spatial smoothness, we add a smoothness constraint $\ell^l_S$, which assumes that the predicted scene flow $SF^l_{\Theta}(p^l_j)$ in a local region $N(p^l_i)$ of $p^l_i$ should be similar to the scene flow at $p^l_i$:
\begin{equation}\label{eq:smoothness}
    \small
    \ell^l_S(SF^l) = \sum_{p^l_i \in P^l}\frac{1}{|N(p^l_i)|}
    \sum_{p^l_j \in N(p^l_i)}\norm{SF^l(p^l_j) - SF^l(p^l_i)}^2_2
\end{equation}
where $|N(p^l_i)|$ is the number of points in the local region $N(p^l_i)$.

\noindent \textbf{Laplacian Regularization} The Laplacian coordinate vector approximates the local shape characteristics of the surface~\cite{sorkine2005laplacian}. The Laplacian coordinate vector $\delta^l(p^l_i)$ is computed as:

\begin{equation}\label{eq:laplacian}\small
    \delta^l(p^l_i) = \frac{1}{|N(p^l_i)|} \sum_{p^l_j\in N(p^l_i)}(p^l_j - p^l_i)
\end{equation}

For scene flow, the warped point cloud $P^l_w$ should have the same Laplacian coordinate vector with the second point cloud $Q^l$ at the same position. Hence, we firstly compute the Laplacian coordinates $\delta^l(p^l_i)$ for each point in second point cloud $Q^l$. Then, we interpolate the Laplacian coordinate of $Q^l$ to obtain the Laplacian coordinate on each point $p^l_w$. We use an inverse distance-based interpolation method similar to Eq.(\ref{eq:interpolation}) to interpolate the Laplacian coordinate $\delta^l$. Let $\delta^l(p^l_w)$ be the Laplacian coordinate of point $p^l_w$ at level $l$, $\delta^l(q^l_{inter})$ be the interpolated Laplacian coordinate from $Q^l$ at the same position as $p^l_w$. 

The Laplacian regularization $\ell^l_L$ is defined as:
\begin{equation}\label{eq:laplacianReg}
    \small
    \ell^l_L(\delta^l(p^l_w), \delta^l(q^l_{inter})) = \sum_{p^l_{w}\in P^l_w}
    \norm{\delta^l(p^l_w) - \delta^l(q^l_{inter})}^2_2
\end{equation}
The overall loss is a weighted sum of all losses across all pyramid levels as:
\begin{equation}\label{eq:self_supervised}
    \small
    \boldsymbol\ell(\Theta) = \sum^{L}_{l=l_0}\alpha_l (\beta_1 \ell^l_C +
    \beta_2 \ell^l_S + \beta_3 \ell^l_L )
\end{equation}
\noindent Where $\alpha_l$ is the factor for pyramid level $l$,  $\beta_1, \beta_2, \beta_3$ are the scale factors for each loss respectively. With the self-supervised loss, our model is able to learn the scene flow from 3D point cloud pairs without any ground truth supervision.

\section{Experiments}
\begin{table}[t]
    \centering
    \caption{\textbf{Evaluation results on the FlyingThings3D and KITTI datasets.} $\mathit{Self}$ means self-supervised, $\mathit{Full}$ means fully-supervised. All approaches are (at least) trained on FlyingThings3D. On KITTI, $\mathit{Self}$ and $\mathit{Full}$ refer to the respective models trained on FlyingThings3D that is directly evaluated on KITTI, while \textit{Self+Self} means the model is firstly trained on FlyingThings3D with self-supervision, then fine-tuned on KITTI  with self-supervision as well. \textit{Full+Self} means the model is trained with full supervision on FlyingThings3D, then fine-tuned on KITTI with self-supervision. ICP~\cite{besl1992method}, FGR~\cite{zhou2016fast}, and CPD~\cite{myronenko2010point} are traditional method that does not require training. Our model outperforms all baselines by a large margin on all metrics}
    \resizebox{\columnwidth}{!}{
    \begin{tabular}{l|l|c|cccc|cc}
        \hline\noalign{\smallskip}
        Dataset &Method         & Sup.  & EPE3D(m)$\downarrow$           & Acc3DS$\uparrow$          & Acc3DR$\uparrow$          & Outliers3D$\downarrow$      & EPE2D(\textit{px})$\downarrow$           & Acc2D$\uparrow$           \\
        \noalign{\smallskip}\hline\noalign{\smallskip}
    \multirow{9}{*}{Flyingthings3D}&ICP(rigid)~\cite{besl1992method} & $\mathit{Self}$& 0.4062          & 0.1614          & 0.3038          & 0.8796          & 23.2280         & 0.2913          \\
        &FGR(rigid)~\cite{zhou2016fast}        &$\mathit{Self}$& 0.4016  & 0.1291   & 0.3461  & 0.8755  & 28.5165  & 0.3037    \\
        &CPD(non-rigid)~\cite{myronenko2010point} &$\mathit{Self}$& 0.4887  & 0.0538   & 0.1694  & 0.9063  & 26.2015  & 0.0966    \\
        &PointPWC-Net   & $\mathit{Self}$& \textbf{0.1213} & \textbf{0.3239} & \textbf{0.6742} & \textbf{0.6878} & \textbf{6.5493} & \textbf{0.4756} \\
        \noalign{\smallskip}\cline{2-9}\noalign{\smallskip}
        &FlowNet3D~\cite{liu2019flownet3d}     &$\mathit{Full}$& 0.1136          & 0.4125          & 0.7706          & 0.6016          & 5.9740          & 0.5692          \\
        &SPLATFlowNet~\cite{su2018splatnet}   &$\mathit{Full}$& 0.1205          & 0.4197          & 0.7180          & 0.6187          & 6.9759          & 0.5512          \\
        &original BCL~\cite{gu2019hplflownet}   &$\mathit{Full}$& 0.1111          & 0.4279          & 0.7551          & 0.6054          & 6.3027          & 0.5669          \\
        &HPLFlowNet~\cite{gu2019hplflownet}     &$\mathit{Full}$& 0.0804          & 0.6144          & 0.8555          & 0.4287 & 4.6723          & 0.6764          \\
        &PointPWC-Net   &$\mathit{Full}$& \textbf{0.0588} & \textbf{0.7379} & \textbf{0.9276} & \textbf{0.3424}  & \textbf{3.2390} & \textbf{0.7994} \\
        \noalign{\smallskip}\hline\noalign{\smallskip}
    \multirow{11}{*}{KITTI}&ICP(rigid)~\cite{besl1992method}        &$\mathit{Self}$& 0.5181          & 0.0669          & 0.1667          & 0.8712          & 27.6752         & 0.1056          \\
        &FGR(rigid)~\cite{zhou2016fast}        &$\mathit{Self}$& 0.4835  & 0.1331   & 0.2851  & 0.7761  & 18.7464  & 0.2876    \\
        &CPD(non-rigid)~\cite{myronenko2010point}        &$\mathit{Self}$& 0.4144  & 0.2058      & 0.4001 & 0.7146 & 27.0583 & 0.1980 \\
        &PointPWC-Net(w/o ft) &$\mathit{Self}$& \textit{0.2549} & \textit{0.2379} & \textit{0.4957} & \textit{0.6863} & \textit{8.9439} & \textit{0.3299} \\
        &PointPWC-Net(w/ ft) &$\mathit{Self + Self}$& \textbf{0.0461} & \textbf{0.7951} & \textbf{0.9538} & \textbf{0.2275} & \textbf{2.0417} & \textbf{0.8645} \\
        \noalign{\smallskip}\cline{2-9}\noalign{\smallskip}
        &FlowNet3D~\cite{liu2019flownet3d}   &$\mathit{Full}$& 0.1767          & 0.3738          & 0.6677          & 0.5271          & 7.2141          & 0.5093          \\
        &SPLATFlowNet~\cite{su2018splatnet} &$\mathit{Full}$& 0.1988          & 0.2174          & 0.5391          & 0.6575          & 8.2306          & 0.4189          \\
        &original BCL~\cite{gu2019hplflownet} &$\mathit{Full}$& 0.1729          & 0.2516          & 0.6011          & 0.6215          & 7.3476          & 0.4411          \\
        &HPLFlowNet~\cite{gu2019hplflownet}   &$\mathit{Full}$& 0.1169 & 0.4783          & 0.7776 & 0.4103          & 4.8055          & 0.5938          \\
        &PointPWC-Net(w/o ft) &$\mathit{Full}$& \textit{0.0694}  & \textit{0.7281} & \textit{0.8884} & \textit{0.2648} & \textit{3.0062} & \textit{0.7673} \\
        &PointPWC-Net(w/ ft) &$\mathit{Full+Self}$& \textbf{0.0430} & \textbf{0.8175} & \textbf{0.9680} & \textbf{0.2072} & \textbf{1.9022} & \textbf{0.8669} \\
        \noalign{\smallskip}\hline
    \end{tabular}
    \label{tab:results}
    }
\end{table}

In this section, we train and evaluate our PointPWC-Net on the FlyingThings3D dataset~\cite{mayer2016large} with the supervised loss and the self-supervised loss, respectively. Then, we evaluate the generalization ability of our model by first applying the model on the real-world KITTI Scene Flow dataset~\cite{Menze2018JPRS,Menze2015ISA} \textit{without any fine-tuning}. Then, with the proposed self-supervised losses, we further fine-tune our pre-trained model on the KITTI dataset to study the best performance we could obtain without supervision. Besides, we also compare the runtime of our model with previous work. Finally, we conduct ablation studies to analyze the contribution of each part of the model and the loss function.

\noindent \textbf{Implementation Details.} We build a 4-level feature pyramid from the input point cloud. The weights $\alpha$ are set to be $\alpha_0=0.02$, $\alpha_1=0.04$, $\alpha_2=0.08$, and $\alpha_3=0.16$, with weight decay $0.0001$.
The scale factor $\beta$ in self-supervised learning are set to be $\beta_1=1.0$, $\beta_2=1.0$, and $\beta_3=0.3$. We train our model starting from a learning rate of $0.001$ and reducing by half every 80 epochs. All the hyperparameters are set using the validation set of FlyingThings3D with 8,192 points in each input point cloud.

\noindent \textbf{Evaluation Metrics.} For fair comparison, we adopt the evaluation metrics that are used in \cite{gu2019hplflownet}. Let $SF_{\Theta}$ denote the predicted scene flow, and $SF_{GT}$ be the ground truth scene flow. The evaluate metrics are computed as follows:

\noindent $\bullet$ \textit{EPE3D(m)}: $\norm{SF_{\Theta}-SF_{GT}}_2$ averaged over each point in meters.

\noindent $\bullet$ \textit{Acc3DS}: the percentage of points with \textit{EPE3D} $< 0.05m$ or relative error $<5\%$.

\noindent $\bullet$ \textit{Acc3DR}: the percentage of points with \textit{EPE3D} $< 0.1m$ or relative error $<10\%$.

\noindent $\bullet$ \textit{Outliers3D}: the percentage of points with \textit{EPE3D}$>0.3m$ or relative error $>10\%$.

\noindent $\bullet$ \textit{EPE2D(px)}: 2D end point error obtained by projecting point clouds back to the image plane.

\noindent $\bullet$ \textit{Acc2D}: the percentage of points whose \textit{EPE2D} $< 3px$ or relative error $<5\%$.

\begin{figure}[t]
    \centering
    \includegraphics[width=0.8\textwidth]{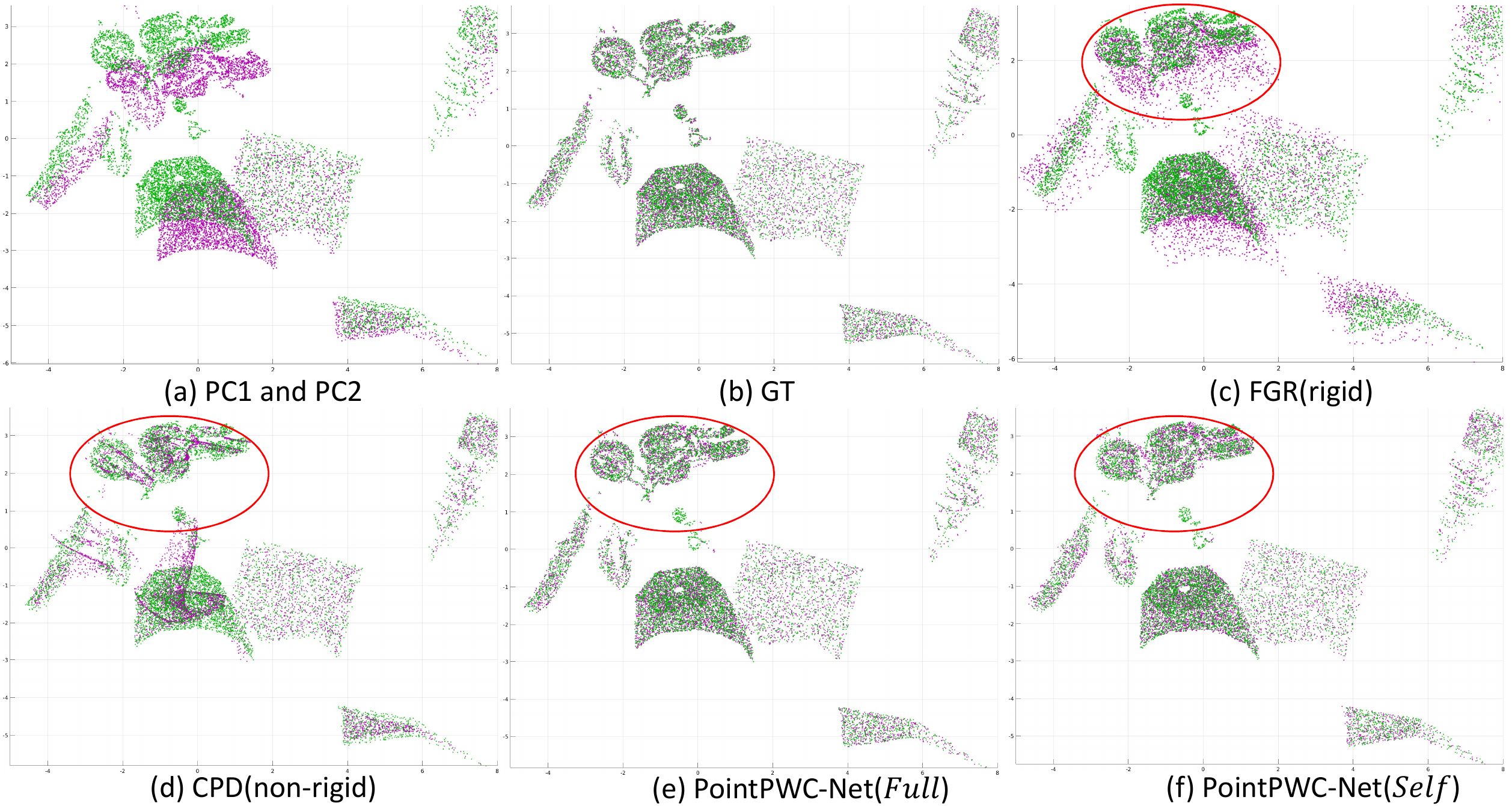}
    \caption{\textbf{Results on the FlyingThings3D dataset.} In (a), 2 point clouds PC1 and PC2 are presented in Magenta and Green, respectively. In (b-f), PC1 is warped to PC2 based on the (computed) scene flow. (b) shows the ground truth; (c) Results from FGR(rigid)~\cite{zhou2016fast}; (d) Results from CPD(non-rigid)~\cite{myronenko2010point}; (e) Results from PointPWC-Net(\textit{Full}); (f) Results from PointPWC-Net(\textit{Self}). Red ellipses indicate locations with significant non-rigid motion. Enlarge images for better view. (Best viewed in color)}
    \label{fig:res_flyingthings3d}
\end{figure}

\subsection{Supervised Learning}
First we conduct experiments with supervised loss. To our knowledge, there is no publicly available large-scale real-world dataset that has scene flow ground truth from point clouds (The input to the KITTI scene flow benchmark is 2D), thus we train our PointPWC-Net on the synthetic Flyingthings3D dataset, following~\cite{gu2019hplflownet}. Then, the pre-trained model is directly evaluated on KITTI Scene Flow 2015 dataset without any fine-tuning. 

\noindent \textbf{Train and Evaluate on FlyingThings3D.}
The FlyingThings3D training dataset includes 19,640 pairs of point clouds, and the evaluation dataset includes 3,824 pairs of point clouds. Our model takes $n=8,192$ points in each point cloud. We first train the model with $\frac{1}{4}$ of the training set(4,910 pairs), and then fine-tune it on the whole training set, to speed up training.

Table~\ref{tab:results} shows the quantitative evaluation results on the Flyingthings3D dataset. 
Our method outperforms all the methods on all metrics by a large margin. 
Comparing to SPLATFlowNet, original BCL, and HPLFlowNet, our method avoids the preprocessing step of building a permutohedral lattice from the input. Besides, our method outperforms HPLFlowNet on \textit{EPE3D} by $\mathit{26.9\%}$. And, we are the only method with \textit{EPE2D} under 4px, which improves over HPLFlowNet by $\mathit{30.7\%}$. See Fig.\ref{fig:res_flyingthings3d}(e) for example results.

\noindent \textbf{Evaluate on KITTI w/o Fine-tune.} To study the generalization ability of our PointPWC-Net, we directly take the model trained using FlyingThings3D and evaluate it on KITTI Scene Flow 2015~\cite{Menze2015ISA,Menze2018JPRS} \textit{without any fine-tuning}. KITTI Scene Flow 2015 consists of 200 training scenes and 200 test scenes. To evaluate our PointPWC-Net, we use  ground truth labels and trace raw point clouds associated with the frames, following \cite{liu2019flownet3d,gu2019hplflownet}. Since no point clouds and ground truth are provided on test set, we evaluate on all 142 scenes in the training set with available point clouds. We remove ground points with height $<0.3m$ following \cite{gu2019hplflownet} for fair comparison with previous methods. 

From Table~\ref{tab:results}, our PointPWC-Net outperforms all the state-of-the-art methods, which demonstrates the generalization ability of our model. For \textit{EPE3D}, our model is the only one below $10cm$, which improves over HPLFlowNet by $\mathit{40.6\%}$. For \textit{Acc3DS}, our method outperforms both FlowNet3D and HPLFlowNet by $\mathit{35.4\%}$ and $\mathit{25.0\%}$ respectively. See Fig.\ref{fig:res_kitti}(e) for example results. 

\begin{table}[t]
    \centering
    \caption{\textbf{Model design.} A learnable cost volume preforms much better than inner product cost volume used in PWC-Net~\cite{sun2018pwc}. Using our cost volume instead of the MLP+Maxpool used in FlowNet3D's flow embedding layer improves performance by $20.6\%$. Compared to no warping, the warping layer improves the performance by $40.2\%$}
    \begin{tabular}{l|c|c}
    \hline\noalign{\smallskip}
    Component                      & Status & EPE3D(m)$\downarrow$ \\
    \noalign{\smallskip}\hline\noalign{\smallskip}
    \multirow{3}{*}{Cost Volume}   & PWC-Net~\cite{sun2018pwc}& 0.0821 \\
                                   & MLP+Maxpool(learnable)~\cite{liu2019flownet3d}& 0.0741 \\
                                   & Ours(learnable) &  \textbf{0.0588} \\
    \noalign{\smallskip}\hline\noalign{\smallskip}
    \multirow{2}{*}{Warping Layer} & w/o      & 0.0984 \\
                                   & w    & \textbf{0.0588} \\
    \noalign{\smallskip}\hline
    \end{tabular}
    \label{tab:modeldesign}
\end{table}

\subsection{Self-supervised Learning}
Acquiring or annotating dense scene flow from real-world 3D point clouds is very expensive, so it would be interesting to evaluate the performance of our self-supervised approach. 
We train our model using the same procedure as in supervised learning, i.e. first train the model with one quarter of the training dataset, then fine-tune with the whole training set. Table~\ref{tab:results} gives the quantitative results on PointPWC-Net with self-supervised learning. We compare our method with ICP(rigid)~\cite{besl1992method}, FGR(rigid)~\cite{zhou2016fast} and CPD(non-rigid)~\cite{myronenko2010point}. Because traditional point registration methods are not trained with ground truth, we can view them as self/un-supervised methods.

\begin{figure}[t]
    \centering
    \includegraphics[width=0.8\textwidth]{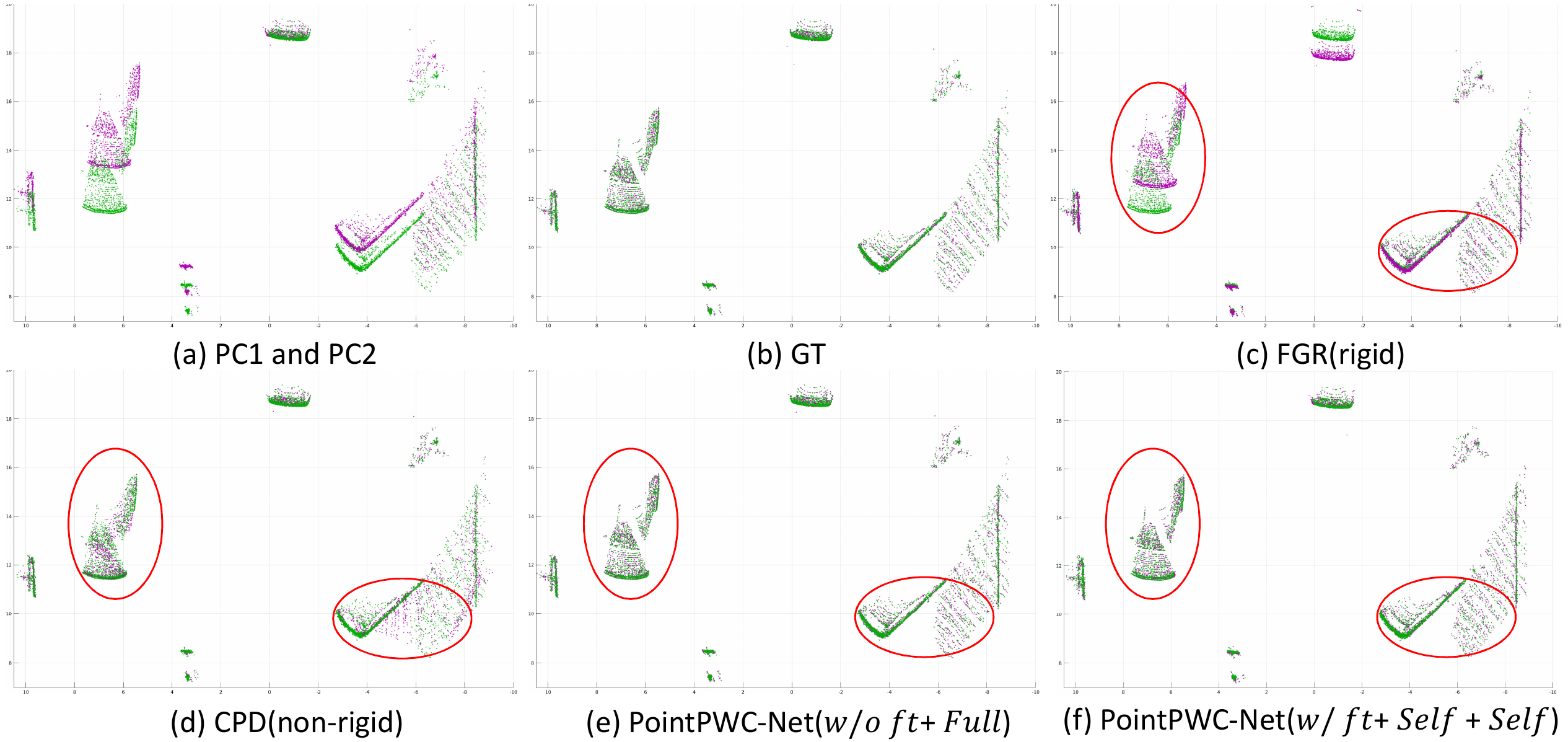}
    \caption{\textbf{Results on the KITTI Scene Flow 2015 dataset.} In (a), 2 point clouds PC1 and PC2 are presented in Magenta and Green, respectively. In (b-f), PC1 is warped to PC2 based on the (computed) scene flow. (b) shows the ground truth; (c) Results from FGR(rigid)~\cite{zhou2016fast}; (d) Results from CPD(non-rigid)~\cite{myronenko2010point}; (e) Results from PointPWC-Net(\textit{w/o ft+Full}) that is trained with supervision on FlyingThings3D, and directly evaluate on KITTI without any fine-tuning; (f) Results from PointPWC-Net(\textit{w/ ft + Self + Self}) which is trained on FlyingThings3D and fine-tuned on KITTI using the proposed self-supervised loss. Red ellipses indicate locations with significant non-rigid motion. Enlarge images for better view. (Best viewed in color)}
    \label{fig:res_kitti}
\end{figure}

\noindent \textbf{Train and Evaluate on FlyingThings3D.}
We can see that our PointPWC-Net outperforms traditional methods on all the metrics with a large margin. See Fig.\ref{fig:res_flyingthings3d}(f) for example results. 

\noindent \textbf{Evaluate on KITTI w/o Fine-tuning.}
Even only trained on FlyingThings3D without ground truth labels, our method can obtain  $\mathit{0.2549m}$ on \textit{EPE3D} on KITTI, which improves over CPD(non-rigid) by $\mathit{38.5\%}$, FGR(rigid) by $\mathit{47.3\%}$, and ICP(rigid) by $\mathit{50.8\%}$. 

\noindent \textbf{Fine-tune on KITTI.} 
With proposed self-supervised loss, we are able to fine-tune the FlyingThings3D trained models on KITTI without using any ground truth. In Table~\ref{tab:results}, the row \textit{PointPWC-Net(w/ ft) Full+Self} and \textit{PointPWC-Net(w/ ft) Self+Self} show the results. \textit{Full+Self} means the model is trained with supervision on FlyingThings3D, then fine-tuned on KITTI without supervision. \textit{Self+Self} means the model is firstly trained on FlyingThings3D, then fine-tuned on KITTI both using self-supervised loss. With KITTI fine-tuning, our PointPWC-Net can achieve \textit{EPE3D} $< 5cm$. Especially, our \textit{PointPWC-Net(w/ ft) Self+Self}, which is fully trained without any ground truth information, achieves similar performance on KITTI as the one that utilized FlyingThings3D ground truth. See Fig.\ref{fig:res_kitti}(f) for example results. 

\subsection{Ablation Study}\label{sec:ablation}
We further conduct ablation studies on model design choices and the self-supervised loss function. On model design, we evaluate the different choices of cost volume layer and removing the warping layer.
On the loss function, we investigate removing the smoothness constraint and Laplacian regularization in the self-supervised learning loss. All models in the ablation studies are trained using FlyingThings3D, and tested on the FlyingThings3D evaluation dataset.

\begin{table}[t]
\centering
\begin{minipage}[t]{0.48\linewidth}
    \centering
    \caption{\textbf{Loss functions.} The Chamfer loss is not enough to estimate a good scene flow. With the smoothness constraint, the scene flow result improves by $38.2\%$. Laplacian regularization also improves slightly}\label{tab:lossfunction}
    \resizebox{\columnwidth}{!}{
        \begin{tabular}{c|c|c|c}
        \hline
        Chamfer & Smoothness & Laplacian & EPE3D(m)$\downarrow$ \\
        \noalign{\smallskip}\hline\noalign{\smallskip}
        \checkmark&  -    &   -   & 0.2112 \\
        \checkmark&\checkmark&  -   &  0.1304 \\
        \checkmark&\checkmark&\checkmark&  \textbf{0.1213} \\
        \noalign{\smallskip}\hline
        \end{tabular}
    }
\end{minipage}
\begin{minipage}[t]{0.48\linewidth}
    \centering
    \caption{\textbf{Runtime.} Average runtime(ms) on Flyingthings3D. The runtime for FlowNet3D and HPLFlowNet is reported from \cite{gu2019hplflownet} on a single Titan V. The runtime for our PointPWC-Net is reported on a single 1080Ti}\label{tab:runtime}
    \resizebox{0.7\columnwidth}{!}{
        \begin{tabular}{l|c}
        \hline\noalign{\smallskip}
        Method      & Runtime(ms)$\downarrow$ \\
        \noalign{\smallskip}\hline\noalign{\smallskip}
        FlowNet3D~\cite{liu2019flownet3d} & 130.8 \\
        HPLFlowNet~\cite{gu2019hplflownet} & 98.4\\
        PointPWC-Net & 117.4 \\
        \noalign{\smallskip}\hline
        \end{tabular}
    }
\end{minipage}%
\end{table}

Tables~\ref{tab:modeldesign}
,~\ref{tab:lossfunction} show the results of the ablation studies. In Table~\ref{tab:modeldesign} we can see that our design of the cost volume obtains significantly better results than the inner product-based cost volume in PWC-Net~\cite{sun2018pwc} and FlowNet3D~\cite{liu2019flownet3d}, and the warping layer is crucial for performance.
In Table~\ref{tab:lossfunction}, we see that both the smoothness constraint and Laplacian regularization improve the performance in self-supervised learning. In Table~\ref{tab:runtime}, we report the runtime of our PointPWC-Net, which is comparable with other deep learning based methods and much faster than traditional ones. 

\section{Conclusion}
To better estimate scene flow directly from 3D point clouds, we proposed a novel learnable cost volume layer along with some auxiliary layers to build a coarse-to-fine deep network, called PointPWC-Net. Because of the fact that real-world ground truth scene flow is hard to acquire, we introduce a loss function that train the PointPWC-Net without supervision. Experiments on the FlyingThings3D and KITTI datasets demonstrates the effectiveness of our PointPWC-Net and the self-supervised loss function, obtaining state-of-the-art results that outperform prior work by a large margin.

\section*{Acknowledgement}
Wenxuan Wu and Li Fuxin were partially supported by the National Science Foundation (NSF) under Project \#1751402, USDA National Institute of Food and Agriculture (USDA-NIFA) under Award 2019-67019-29462, as well as by the   Defense   Advanced   Research   Projects   Agency (DARPA)   under   Contract   No. N66001-17-12-4030 and N66001-19-2-4035. Any  opinions, findings and conclusions or recommendations expressed in this material are those of the author(s) and do not necessarily reflect the views of the funding agencies.

\section*{Appendix}
In the appendix, we provide more details on the network structures of the feature pyramid network and the scene flow predictor in Sec.\ref{sec:network}. In Sec.\ref{sec:experiments}, we conduct additional ablation experiments. Besides, we also provide more visualization of scene flow results on the KITTI Scene Flow 2015~\cite{Menze2018JPRS,Menze2015ISA} dataset in Sec.\ref{sec:visualization}. Finally, we analyze the typical error types of PointPWC-Net on KITTI in Sec.~\ref{sec:errortypes}. 

\subsection{Network Details}\label{sec:network}

\begin{figure}
    \centering
    \includegraphics[width=0.8\textwidth]{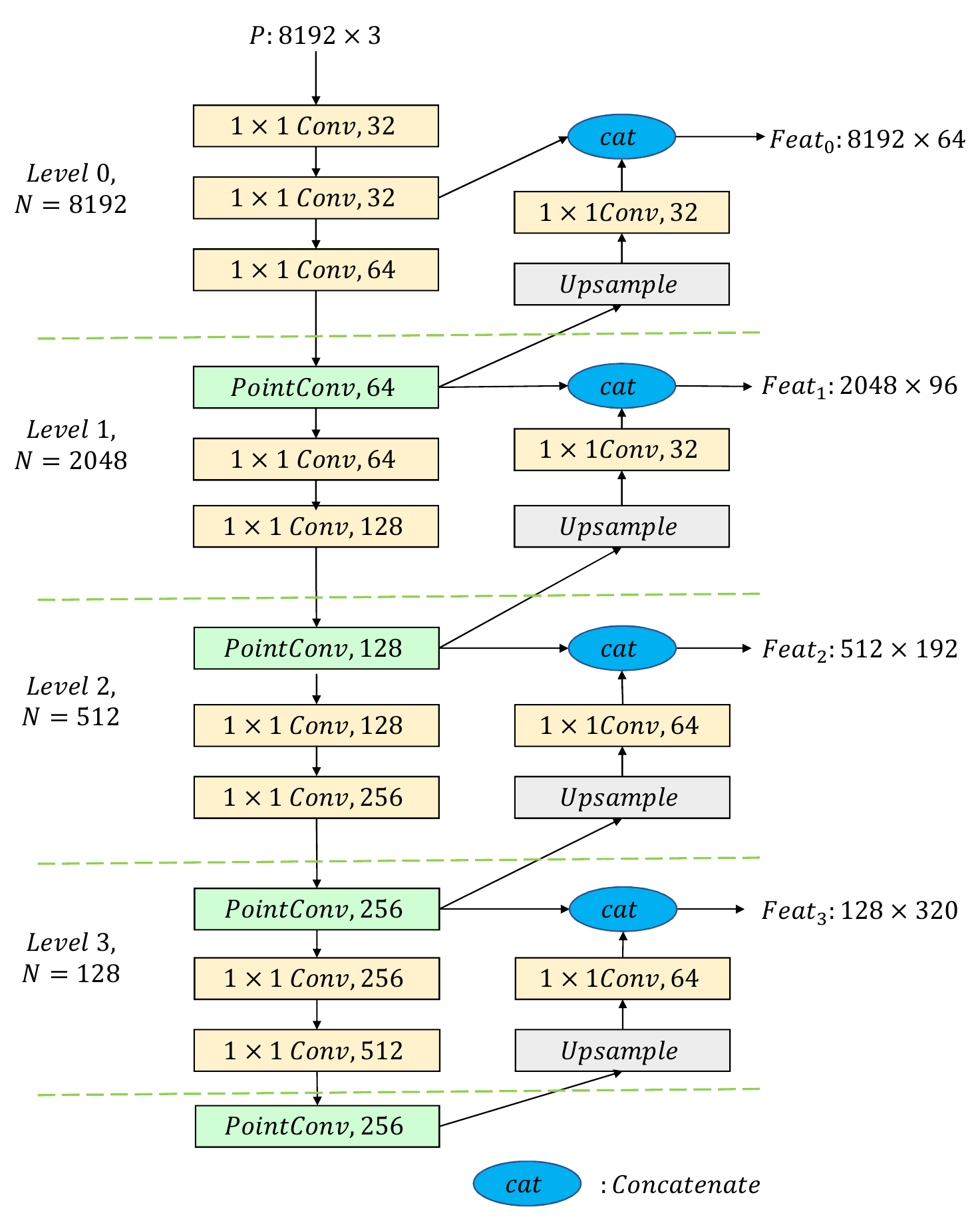}
    \caption{\textbf{Feature Pyramid Network.} The first point cloud and the second point cloud are encoded using the same network with shared weights. For each point cloud, we use PointConv~\cite{wu2019pointconv} to convolve and downsample by factor of 4. The $1 \times 1 Conv$s are used to increase the representation power and efficiency. The final feature of level $l$ is concatenated with the upsampled feature from level $l + 1$, which contains feature with a larger receptive field. }
    \label{fig:featurepyramid}
\end{figure}

\begin{figure}
    \centering
    \includegraphics[width=0.8\textwidth]{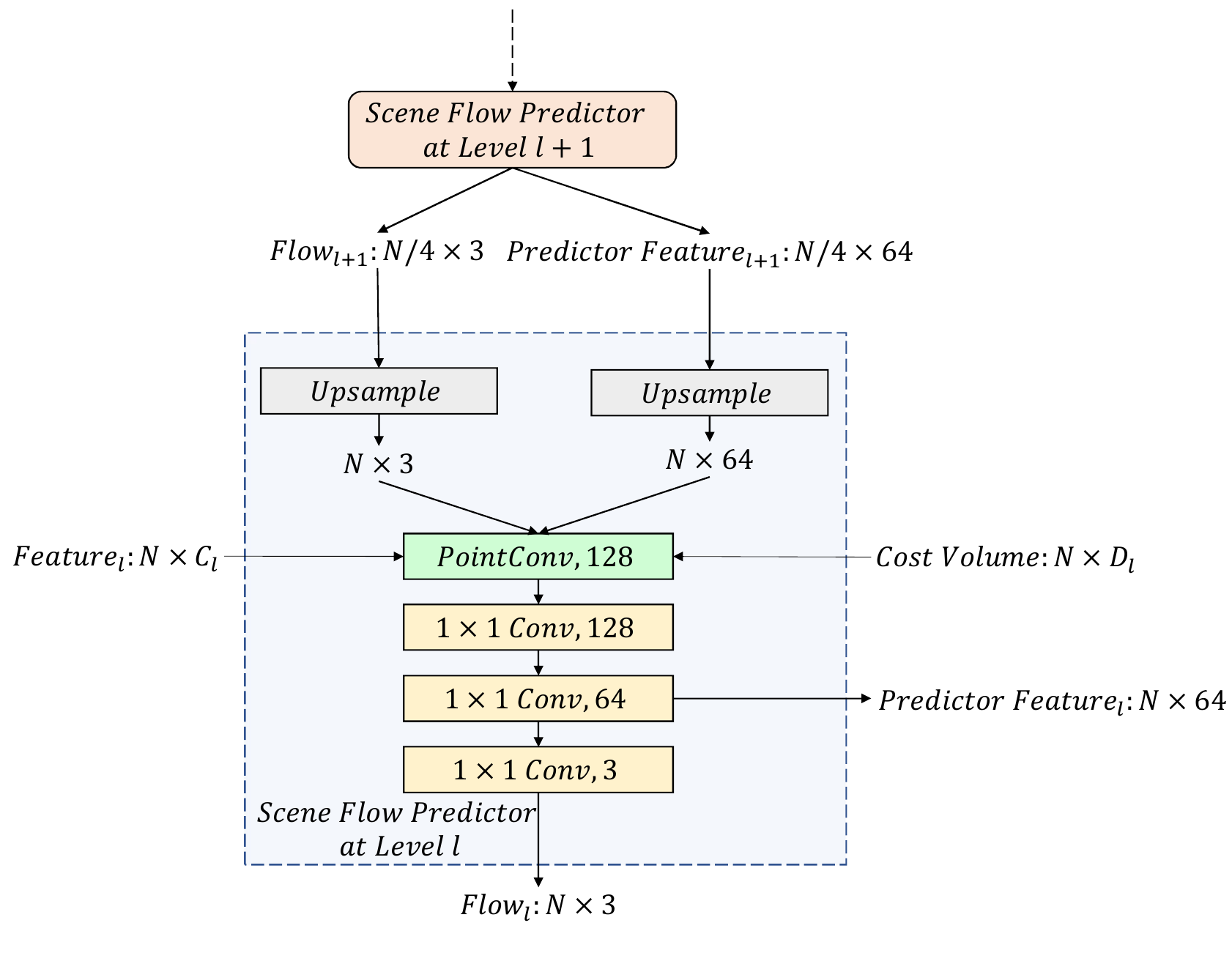}
    \caption{\textbf{Scene Flow Predictor.} The scene flow predictor takes the feature from the first point cloud, the cost volume, the upsampled flow from previous layer, and the upsampled feature of the second last layer from previous level's scene flow predictor as input. The output is the estimated flow in current level and the feature in the second last layer.}
    \label{fig:sceneflowpredictor}
\end{figure}

Fig.\ref{fig:featurepyramid} shows the architecture for feature pyramid network. Fig.\ref{fig:sceneflowpredictor} shows the scene flow predictor network. At the most coarse level, the scene flow predictor only takes the feature from the first point cloud and the cost volume as input. At the rest of the level, the scene flow predictor takes the feature from the first point cloud, the cost volume, the upsampled flow from previous layer, and the upsampled feature of the second last layer from previous level's scene flow predictor as input.

\subsection{Additional Experiments}\label{sec:experiments}

\noindent \textbf{Concatenated coarser features.} In feature pyramid network, we concatenate the feature on level $l$ with the upsampled feature from level $l + 1$, which gives a larger receptive field. In scene flow predictor, we take the feature from the second last layer of scene flow predictor from level $l + 1$ as an input to the scene flow predictor in level $l$. In Table \ref{tab:features}, we show the effectiveness of these features. Both features can improve our scene flow results significantly.

\noindent \textbf{Runtime breakdown.} In order to study the runtime of each critical part of our model, we include the runtime breakdown in Table.~\ref{tab:runtime_breakdown}. 

\begin{table}[t]
    \centering
    \caption{\textbf{Concatenated coarser features.}
    The upsampled feature at level $l$ is the feature upsampled from the level $l + 1$ followed by a $1 \times 1 Conv$. The predictor feature for level $l$ is upsampled from the second last layer of the scene flow predictor in level $l + 1$. Adding just the upsampled feature improves the performance by $9.5\%$, and then with both the performance improves by $34.8\%$}
    \label{tab:features}
    \begin{tabular}{c|c|c}
    \hline
    Upsampled Feature & Predictor Feature & EPE3D$\downarrow$ \\
    \noalign{\smallskip}\hline\noalign{\smallskip}
      -  &   -   & 0.0902 \\
    \checkmark&  -  &  0.0816 \\
    \checkmark&\checkmark&  \textbf{0.0588} \\
    \noalign{\smallskip}\hline
    \end{tabular}
\end{table}

\begin{table}[t]
    \centering
    \caption{\textbf{Runtime breakdown.} The runtime of each critical component in PointPWC-Net}
    \begin{tabular}{l|cccc}
    \hline\noalign{\smallskip}
    \multirow{2}{*}{Component} & Feature & Cost   & Upsample & Scene Flow \\
                      & Pyramid & Volume & Warping   & Predictor \\
    \noalign{\smallskip}\hline\noalign{\smallskip}
    Runtime(ms)       & 43.2    & 22.7   & 24.5      & 27.0       \\
    \noalign{\smallskip}\hline
    \end{tabular}
    \label{tab:runtime_breakdown}
\end{table}

\subsection{More Visual Results on KITTI}\label{sec:visualization}

\begin{figure}
    \centering
    \includegraphics[width=0.95\textwidth]{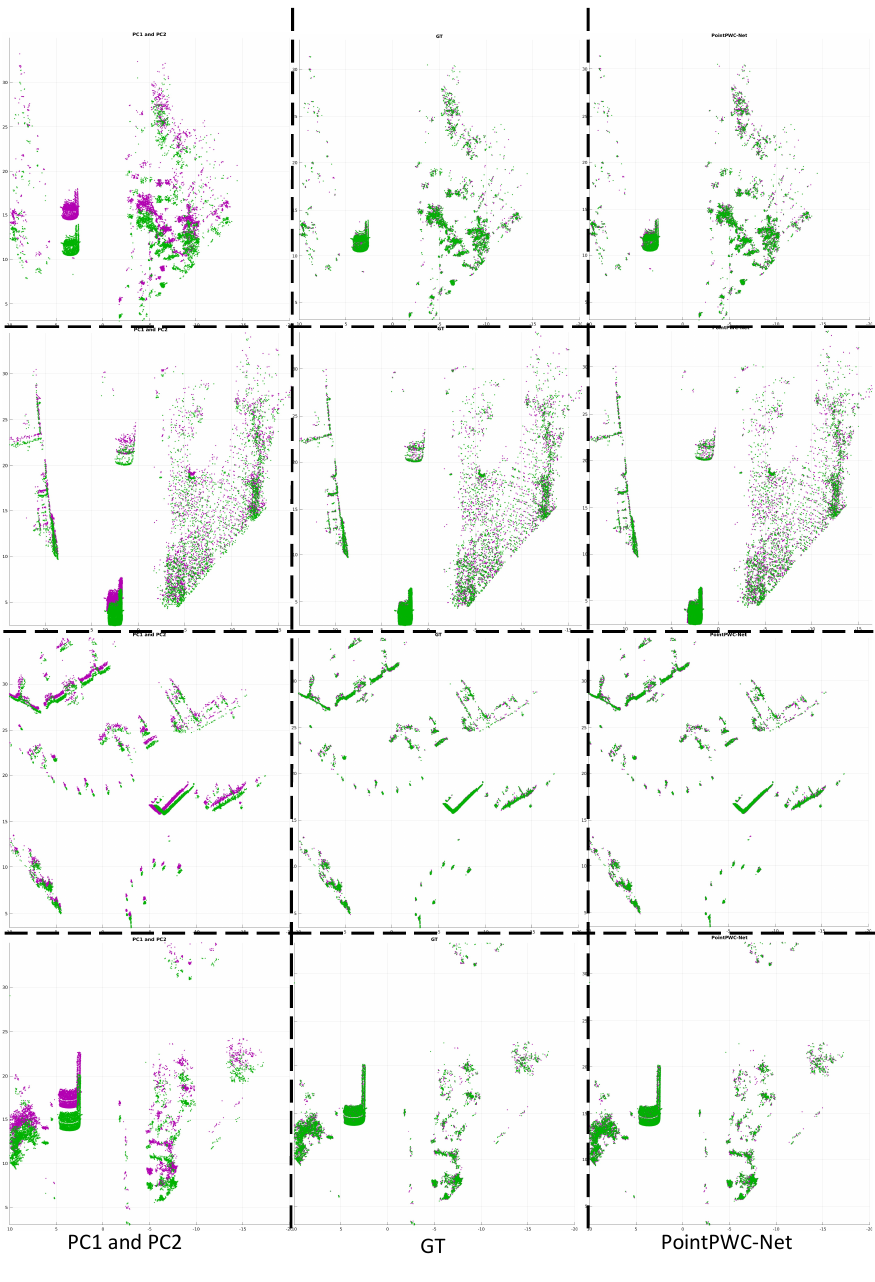}
    \vspace{-0.2in}
    \caption{\textbf{Scene Flow Results on KITTI Scene Flow 2015}. In (a), 2 point clouds PC1 and PC2 are presented in Magenta and Green, respectively. In (b) and (c), PC1 is warped to PC2 based on the (computed) scene flow. (b) shows the ground truth; (c) Results from PointPWC-Net that is trained with supervision on FlyingThings3D, and directly evaluate on KITTI without any fine-tune. Enlarge images for better view. (Best viewed in color)}
    \label{fig:kitti_self_supp}
\end{figure}

Fig.\ref{fig:kitti_self_supp} provide more visualization on KITTI dataset of scene flow results by PointPWC-Net. In Fig.\ref{fig:kitti_self_supp}, the model is trained with self-supervised loss on FlyingThings3D~\cite{mayer2016large}, and directly tested on KITTI Scene Flow 2015~\cite{Menze2018JPRS,Menze2015ISA} without any finetune. From Fig.\ref{fig:kitti_self_supp}, we can see that our model can recover scene flow not only for the rigid objects, such as cars, but also for the non-rigid objects, such as bushes, trees, etc.

\subsection{Typical Error Types}\label{sec:errortypes}

\begin{figure}
    \centering
    \includegraphics[width=0.8\textwidth]{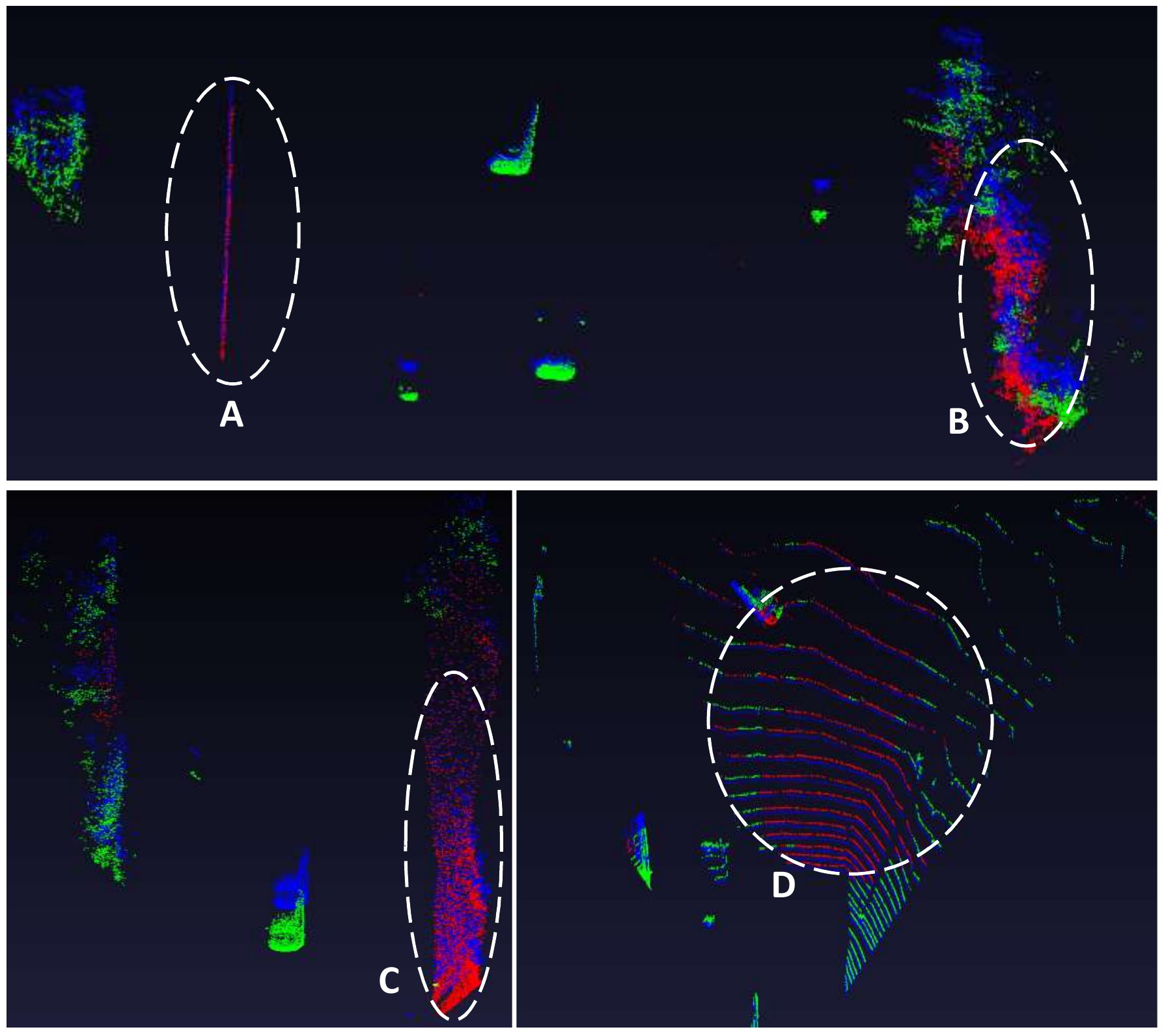}
    \caption{\textbf{Typical error types for KITTI.} The blue points are from the first point cloud $P$. The green points are the warped points $P_w=P + SF$ according to the correctly predicted flow. The ``correctness'' is measured by Acc3DR. The red points are wrongly predicted. $\mathbf{A}$ and $\mathbf{C}$ are the ambiguity in 3D point clouds, which are straight lines or plane walls.
    $\mathbf{B}$ is the messy bushes, whose features do not have strong correspondences. $\mathbf{D}$ is the case when the ground points are not removed cleanly.}
    \label{fig:fail_case}
\end{figure}

We summaries three typical error types of our PointPWC-Net for KITTI dataset. To visualize errors, we use blue, green and red to represent the first point cloud, the warped points which are correctly predicted, and the wrongly predicted points, respectively, as shown in Fig.~\ref{fig:fail_case}. The first error type is when the object is a straight line or a plane. In this case, it is hard for the network to construct a cost volume with strong discernment, as shown in Fig.\ref{fig:fail_case}~$\mathbf{A}$ and $\mathbf{C}$. The second one is that it is hard to find good correspondences between consecutive frames due to the strong deformation of local shapes, as shown in Fig.\ref{fig:fail_case}~$\mathbf{B}$. The third case is that the ground points are not removed properly, as shown in Fig.\ref{fig:fail_case}~$\mathbf{D}$. By using a better ground removal strategy, we can further improve our results.

\clearpage
%
%

\end{document}